\documentclass[preprint,authoryear,10pt,3p, twocolumn]{elsarticle}

\usepackage{framed,multirow}
\usepackage{amsmath}
\usepackage{tabularx,booktabs}
\usepackage{balance}
\usepackage[draft]{hyperref}
\usepackage{breqn}

\usepackage{amssymb}
\usepackage{latexsym}
\usepackage{multirow}
\usepackage{gensymb}
\usepackage[ruled,vlined,linesnumbered]{algorithm2e}

\usepackage{url}
\usepackage{xcolor}

\usepackage{hyperref}

\journal{Medical Image Analysis}

\begin{document}


\begin{frontmatter}

\title{Fully Automated Left Atrium Segmentation from Anatomical Cine Long-axis MRI Sequences using Deep Convolutional Neural Network with Unscented Kalman Filter}%

\author[1,2,3]{Xiaoran Zhang}
\cortext[cor1]{Corresponding author: Kumaradevan Punithakumar (Email: punithak@ualberta.ca)}
\author[2,3]{Michelle Noga}
\author[2,3]{David Glynn Martin}
\author[2,3,4]{Kumaradevan Punithakumar\corref{cor1}}

\address[1]{Department of Electrical and Computer Engineering, University of California, Los Angeles}
\address[2]{Department of Radiology and Diagnostic Imaging, University of Alberta, Edmonton, Canada}
\address[3]{Servier Virtual Cardiac Centre, Mazankowski Alberta Heart Institute, Edmonton, Canada}
\address[4]{Department of Computing Science, University of Alberta, Edmonton, Canada}


\begin{abstract}
This study proposes a fully automated approach for the left atrial segmentation from routine cine long-axis cardiac magnetic resonance image sequences using deep convolutional neural networks and Bayesian filtering. The proposed approach consists of a classification network that automatically detects the type of long-axis sequence and three different convolutional neural network models followed by unscented Kalman filtering (UKF) that delineates the left atrium. Instead of training and predicting all long-axis sequence types together, the proposed approach first identifies the image sequence type as to 2, 3 and 4 chamber views, and then performs prediction based on neural nets trained for that particular sequence type. The datasets were acquired retrospectively and ground truth manual segmentation was provided by an expert radiologist. In addition to neural net based classification and segmentation, another neural net is trained and utilized to select image sequences for further processing using UKF  to impose temporal consistency over cardiac cycle. A cyclic dynamic model with time-varying angular frequency is introduced in UKF to characterize the variations in cardiac motion during image scanning. The proposed approach was trained and evaluated separately with varying amount of training data with images acquired from 20, 40, 60 and 80 patients. Evaluations over 1515 images with equal number of images from each chamber group acquired from an additional 20 patients demonstrated that the proposed model outperformed state-of-the-art and yielded a mean Dice coefficient value of 94.1\%, 93.7\% and 90.1\% for 2, 3 and 4-chamber sequences, respectively, when trained with datasets from 80 patients.
\end{abstract}


\end{frontmatter}


\section{Introduction}
The assessment of left atrial function is becoming increasingly important due to its role in prognosis and risk stratification of several cardiovascular diseases including cardiomyopathy, ischemic heart disease and valvular heart disease \citep{Hoit2014,Kowallick2014,tobon2015benchmark}. While both the Computed Tomography (CT) and Magnetic Resonance Imaging (MRI) could be used for the anatomical assessment, the functional assessment of the left atrium is often performed using cine MRI or echocardiography. The advances in cardiac MRI technology have led to the generation of a large amount of imaging data with an increasingly high level of quality. However, segmenting a large number of left atrial images in cardiac MRI over the entire cardiac cycle is a tedious and complex task for clinicians, who have to manually extract important information \citep{despotovic2015mri}. The manual analysis is often time-consuming and subject to inter- and intra-operator variability. Due to these difficulties, there is an increasing demand for computerized methods to shorten segmentation time and improve the accuracy of cardiac diagnosis and treatment.

A number of cardiac image segmentation methods using CT or MRI have been proposed in the literature to improve the accuracy of the cardiac chamber delineations \citep{Peng2016,zheng2008four,budge2008analysis,kaus2004automated,radau2009evaluation}. However, most of these methods focus on automated delineation of the left ventricle only \citep{Peng2016,petitjean2011review,zreik2016automatic}.  The left atrium is more difficult to segment \citep{tobon2015benchmark}. The difficulties arise from following reasons \citep{zhu2012automatic}: 1) The wall of the left atrium is relatively thin compared to the left ventricle; 2) Boundaries are not clearly defined when the blood pool of the left atrium goes into pulmonary veins; and 3) The shape variability of the left atrium is large between different patients. There are a few methods proposed in the literature to achieve automated left atrial segmentation using model based methods \citep{ecabert2008automatic,ecabert2011segmentation,ordas2007statistical}, which rely on static images obtained using CT imaging \citep{hunold2003radiation}. In comparison to CT imaging, MRI produces higher temporal resolution, without the use of ionizing radiation or intravenous contrast media, and cine MRI allows for assessment of motion over the entire cardiac cycle \citep{parry2005advantages}.

A few notable exceptions to studies proposed for the segmentation of the left atrium from MRI images include a method based on traditional image analysis techniques such as salient feature and contour evolution by \citet{zhu2012automatic} and deep learning based method by \citet{mortazi2017cardiacnet}. In addition, a public segmentation challenge to delineate the chamber from gadolinium-enhanced 3D MRI sequences was hosted at the Statistical Atlases and Computational Models of the Heart (STACOM) workshop in the Medical Image Computing and Computer Assisted Intervention (MICCAI) conference in 2018 \citep{pop2019statistical}. More than 15 research teams participated in the challenge and a V-net \citep{Milletari2016} based  convolutional neural network approach by \citet{Xia2019} yielded the best performance. However, these methods rely on 3D MRI scans of a single point in the cardiac phase which may not be suited for the functional assessment of the left atrium \citep{zhang_fully_2018}.

This study proposes a fully automated left atrial segmentation approach for long-axis cine MRI sequences. First, a classification network is introduced to automatically detect the input sequence into three groups according to the number of cardiac chambers visible in the images, \emph{i.e.,} the network will classify the sequence as 2-chamber, 3-chamber or 4-chamber long-axis cine MRI. Then, a U-net \citep{ronneberger2015u} based model which was trained with 5-fold cross-validation for the detected sequence is applied. In order to improve the temporal consistency of the delineations, a second classification network is trained to detect image sequences which tend to yield low Dice coefficient in comparison to manual contours and the unscented Kalman filter (UKF) with a periodic dynamic model is utilized to improve the accuracy of the selected sequences. Evaluations over 1515 images acquired from 20 patients in comparison to expert manual delineation for each chamber group demonstrated that the proposed method yielded significantly better results than U-net alone. Moreover, this study empirically assesses the impact of the number f training samples on the network performance by comparing the results of four different scenarios where the size of training sets are 20, 40, 60 and 80 patients.

\section{Method}
\begin{figure*}[htbp]
\centerline{\includegraphics[scale=0.50]{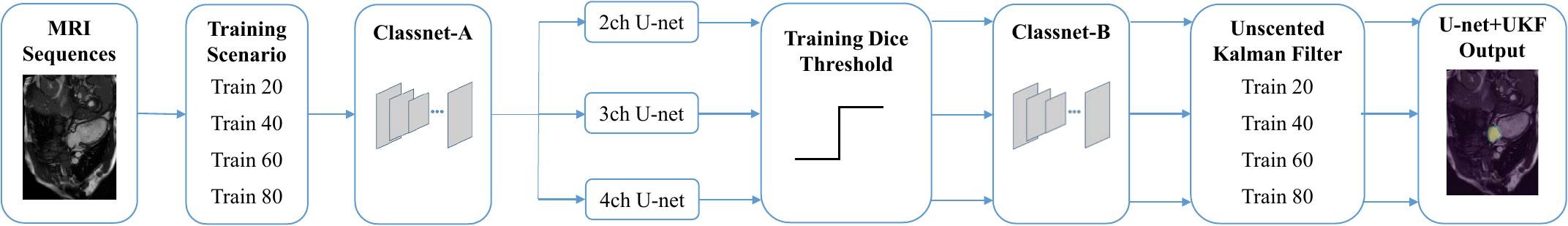}}
\caption{Overall system diagram showing the components of the proposed method as well as different test scenarios of this study to empirically assess the impact of the size of the training set on the performance of the U-net and the proposed U-net+UKF algorithms. Four different test scenarios were created. The accuracy of the algorithms were evaluated using overlap and distance metrics in comparison to expert manual delineation of the left atrium from 2, 3 and 4-chamber long-axis MRI sequences.}
\label{fig1}
\end{figure*}

The objective of the proposed method is to automatically delineate the left atrium from cine long-axis 2D MRI sequences correspond to 2-chamber, 3-chamber and 4-chamber views that are acquired in routine clinical scans. The overall system diagram in Fig.~\ref{fig1} depicts the components of the proposed method which integrates a classification network to detect chamber type, three U-net based frameworks for semantic segmentation followed by another classification network to detect results with low accuracy and unscented Kalman filter to improve temporal consistency. Both proposed classifying networks are based on convolutional neural networks and shortened as \textit{Classnet-A} and \textit{Classnet-B} in rest of the paper. Each input image is resized to $256\times256$ using zero-padding scheme along the image borders.
Instead of training blindly with all image data, three separate U-net based neural nets corresponding to 2, 3 and 4-chamber sequences were constructed and trained after classifying the sequence using the \textit{Classnet-A}. The network architecture of the proposed \textit{Classnet-A} is given in Fig.~\ref{fig:ClassnetA}. The standard U-net model is limited when processing image sequences as the predictions are made independently where the temporal consistency is not maintained. In order to improve the accuracy, the unscented Kalman filter to exploit time-serial aspect of cardiac motion in the MRI sequence data. The second classifier neural net, \textit{Classnet-B}, is trained using to the results from the U-net in comparison to expert manual ground truth to detect image sequences that require further processing. The unscented Kalman filter is then applied as a post-processing step for image sequences for the selected sequences.

\subsection{Classnet Architecture}

\begin{figure*}[htbp]
    \centering
    \includegraphics[scale=0.6]{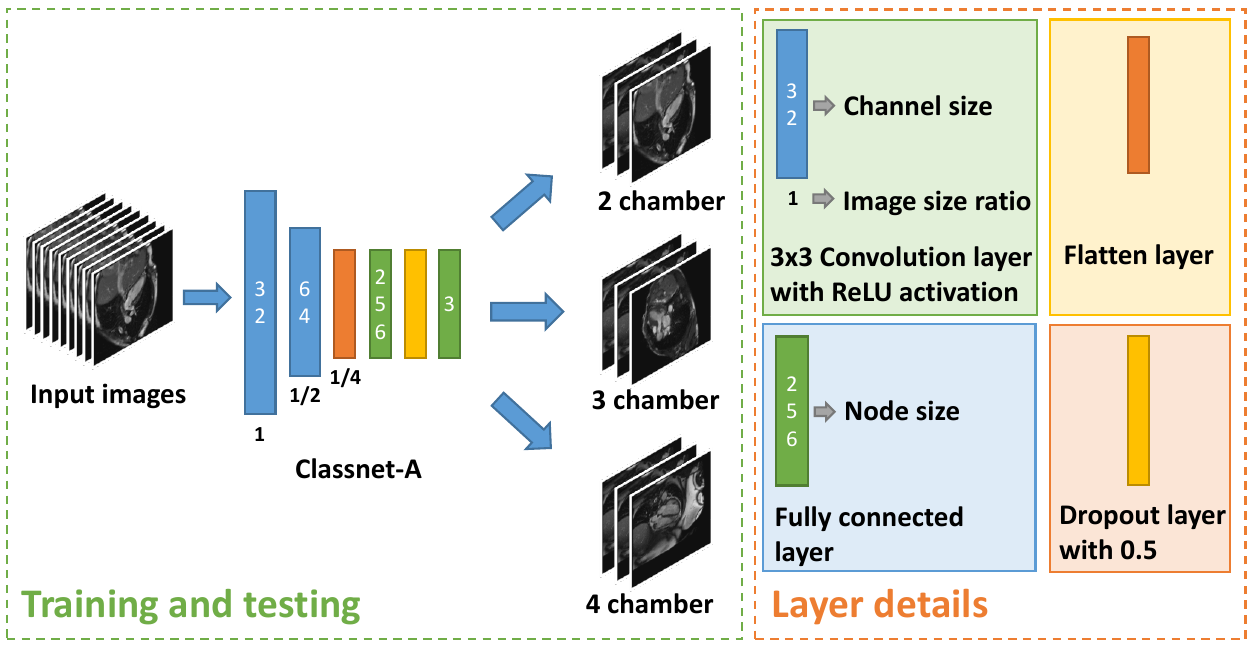}
    \caption{\textit{Classnet-A} architecture proposed in this paper to automatically detect the input image sequence and classify the sequence type. The input images are resized to $256\times256$. The image size is changed by applying a $2\times2$ maxpooling layer. A fully connected block with softmax activation is applied in the final layer to obtain output class.}
    \label{fig:ClassnetA}
\end{figure*}

As shown in Fig. \ref{fig:ClassnetA}, the image is processed by a $3\times3\times32$ convolutional layer with ReLU activation function followed by a $2\times2$ max pooling layer. Another $3\times3\times64$ convolutional and $2\times2$ max pooling layers are added afterwards. The output of the max pooling layer is connected to a fully connected block with drop out (0.5). The final fully connected layer outputs the classification result of the chamber group after a softmax activation. 

\subsection{U-net Architecture}

\begin{figure*}
    \centering
    \includegraphics[scale=0.6]{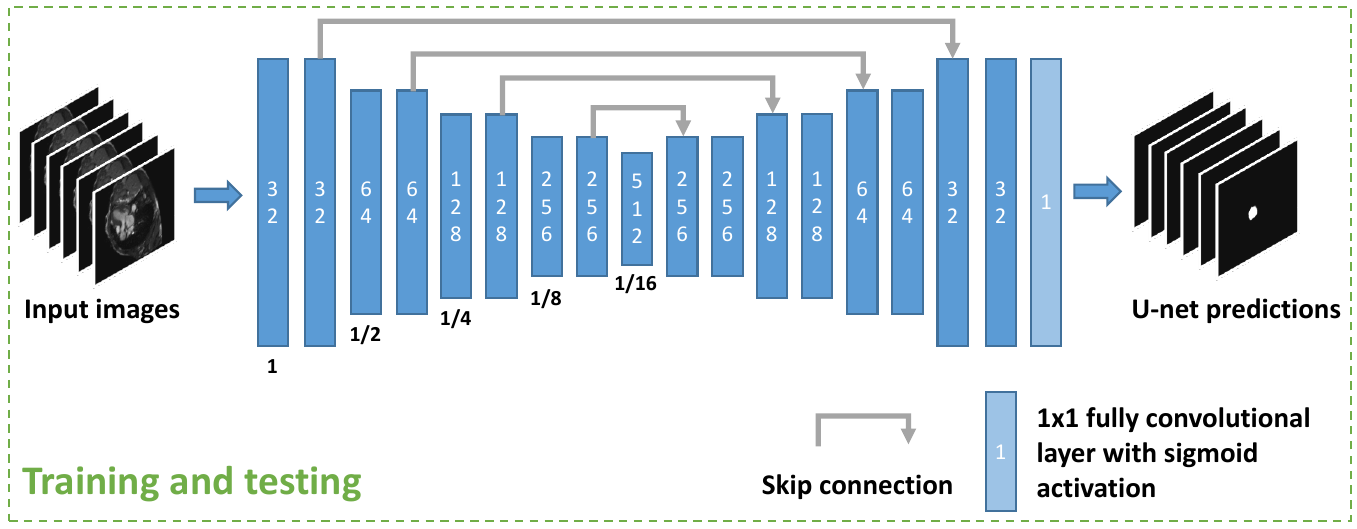}
    \caption{U-net architecture for each chamber group. The convolution layer follows the same layer details in Fig.~\ref{fig:ClassnetA}. Final output is a $256\times256$ image mask corresponding to the left atrial semantic segmentation.}
    \label{fig:Unet}
\end{figure*}

Three U-net based models are applied exclusively after classifying the input images using \textit{Classnet-A}. An example U-net model used in this study is depicted in Fig.~\ref{fig:Unet}. Similar to the original U-net architecture \citep{ronneberger2015u}, each U-net based model contains convolutional with ReLU activation function, max pooling, skip connection and upsampling operations. It then connects to a fully convolutional layer with sigmoid activation function to obtain output masks. The labelled masks for training and testing are both scaled to $[0,1]$. The loss function is chosen as the negative of Dice metric given in \eqref{eq1}. The input images are normalized with respect to the 1$^{\textrm {st}}$ and 99$^{\textrm{th}}$ percentile of pixel intensity values of each image.

\subsection{Image Selection for Post-processing}
\begin{figure*}[htbp]
    \centering
    \includegraphics[scale=0.54]{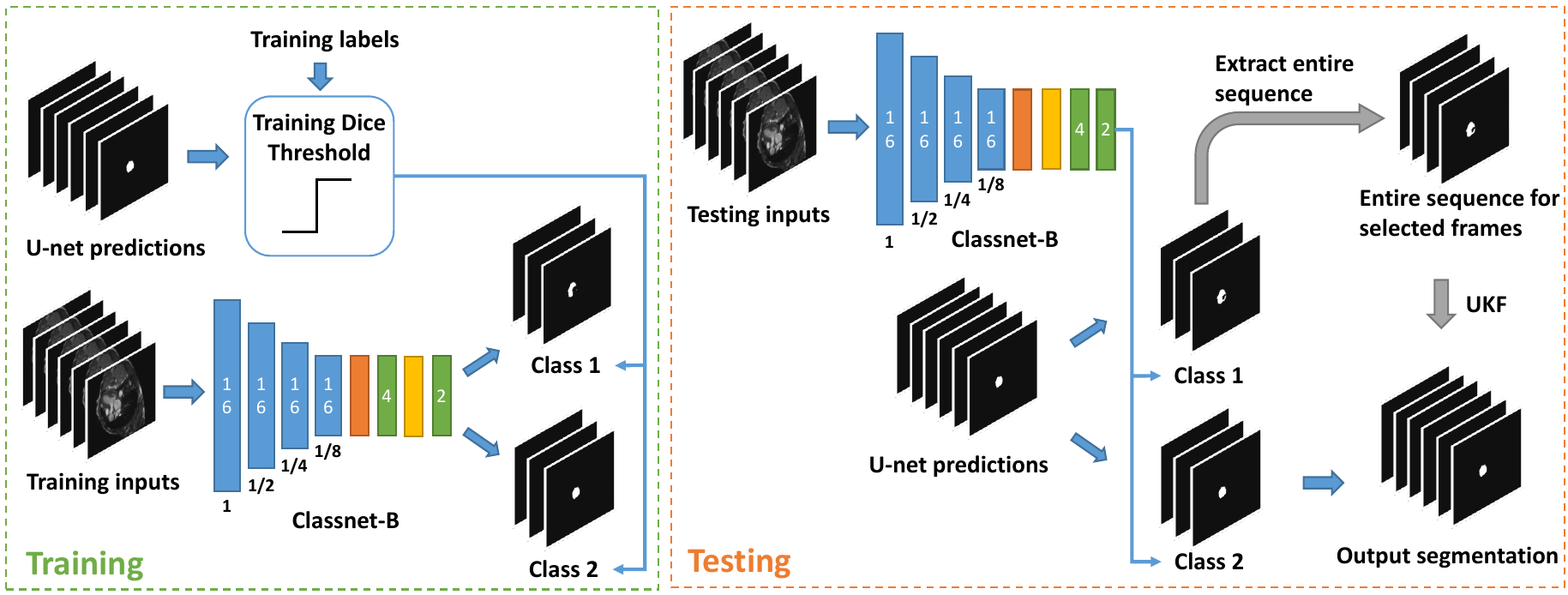}
    \caption{Visualization of the training and application of the \textit{Classnet-B} neural net of the proposed method. The algorithm was trained using the input MRI images along with the classification labels obtained based on the Dice score values of the U-net predictions where a threshold of 0.9 was used in deciding the class labels. During application or testing stage, the \textit{Classnet-B} predictions for the input MRI images were used in deciding the application of the UKF for the U-net predictions.}
    \label{fig:ClassnetB}
\end{figure*}
After obtaining semantic segmentation results from U-net based models, the \textit{Classnet-B} is applied to select the images for post-processing. The training and testing process is shown in Fig.~\ref{fig:ClassnetB}. The layer details follow the same notation defined in Fig.~\ref{fig:ClassnetA}. The \textit{Classnet-B} neural net is trained using input MRI images from the training set with binary annotations consisting of Class 1 and Class 2 where the classes are defined based on the U-net prediction accuracy. The training set images which yield a Dice score less than the threshold 0.9 are designated as Class 1 and the rest of the images in the set that yield a Dice score of 0.9 and above are designated as Class 2.
The rationale for the selection is that images with certain degraded quality lead to poor prediction performance using U-net. The \textit{Classnet-B} was trained using the comparative results by U-net predictions and expert manual ground truth for the training data sets. Upon training, the \textit{Classnet-B} is applied to the testing set to identify image frames that will likely yield diminished accuracy (Dice score $< 0.9$) with U-net based predictions, without relying on ground truth labels. If any of the image frames in a sequence is classified as Class 1 by the \textit{Classnet-B}, the entire sequence is selected for post-processing with the UKF.

\subsection{Unscented Kalman Filter}

\begin{figure*}[htbp]
    \centering
    \includegraphics[scale=0.6]{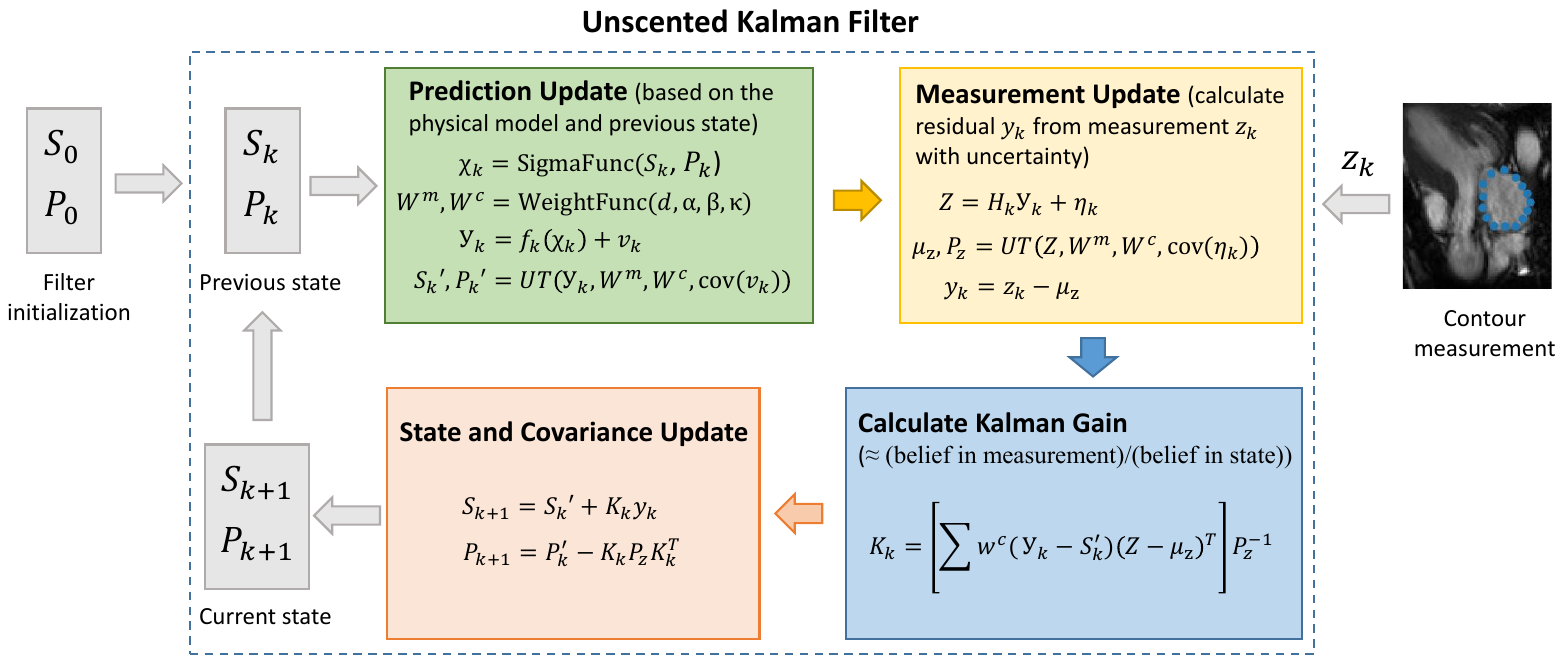}
    \caption{Visualization of the UKF post-processing process for each sample contour point over the cardiac cycle. The objective of the UKF post-processing is to enforce temporally consistent segmentation results by considering the image frames in the entire cardiac sequence in contrast to the processing by U-net which segments each frame independently. UT denotes unscented transform. The proposed method uses a state variable with dimension $d=7$ and unscented transformation parameters $\alpha=0.1, \beta=2, \kappa=-1$.}
    \label{fig:ukf_mainfig}
\end{figure*}

\begin{algorithm*}[h]
    \caption{Unscented Kalman filter for enforcing temporal consistency}\label{alg_ukf}
    \textbf{Input}: U-net prediction after uniform sampling with shape $N\times K\times 2$. $K$ is the total number of frames in each cardiac sequence and $N$ is the number of samples in contour.\\
    \textbf{For} contour point $i=1,2,3,...,N$ \textbf{do}\\
    \qquad \textbf{Initialize:} $S_1^i=[\hat{\dot{x_1}} \; \hat{x_1} \; \hat{\bar{x_1}} \; \hat{\dot{y_1}} \; \hat{y_1} \; \hat{\bar{y_1}} \; \omega_1]^T$ in \eqref{eq24}-\eqref{eq27}, $P_1^i$ in \eqref{eq28} and \eqref{eq29}, $Q_1^i$ in \eqref{eq20}, $R_1^i$ in \eqref{eq23}.\\
    \qquad\qquad \textbf{For} time frame $k=1,2,3,...,K$ \textbf{do}:\\
    
    \qquad\qquad\qquad \textbf{Prediction update:}\\
    
    \qquad\qquad\qquad $\;$ $\chi_k^i=\text{SigmaFunction}(S_k^i,P_k^i)$\\
    \qquad\qquad\qquad $\;$ $W^m,W^c=\text{WeightFunction}(d,\alpha,\beta,\kappa)$\\
    \qquad\qquad\qquad $\;$ $\mathcal{Y}_k^i=f_k(\chi_k^i)+v_k$\\
    \qquad\qquad\qquad $\;$ $S_k^{i'},P_k^{i'}=\text{UnscentedTransform}(\mathcal{Y}_k^i,W^m,W^c,Q_1^i)$\\
    
    \qquad\qquad\qquad \textbf{Measurement update:}\\
    \qquad\qquad\qquad $\;$ $\mathcal{Z}=H_k\mathcal{Y}_k^i+\eta_k$ \\
    \qquad\qquad\qquad $\;$ $\mu_z^i,P_z^i=\text{UnscentedTransform}(\mathcal{Z},W^m,W^c,R_1^i)$\\
    \qquad\qquad\qquad $\;$ $y_k^i=z_k^i-\mu_z^i$\\
    
    \qquad\qquad\qquad \textbf{Calculate Kalman gain:}\\
    \qquad\qquad\qquad $\;$ $K_k^i=[\sum w^c(\mathcal{Y}_k^i-S_k^{i'})(\mathcal{Z}-\mu_z^i)^T](P_z^i)^{-1}$\\
    
    \qquad\qquad\qquad \textbf{State and covariance update:}\\
    \qquad\qquad\qquad $\;$ $S_{k+1}^i=S_k^{i'}+K_k^iy_k^i$ \\
    \qquad\qquad\qquad $\;$ $P_{k+1}^i=P_k^{i'}-K_k^iP_z^i(K_k^i)^T$
    
    \textbf{Output:} Filtered contour with shape $N\times K\times 2$ and improved temporal consistency.
\end{algorithm*}

The UKF is utilized for sequences which were detected by the \textit{Classnet-B} that are likely to yield diminished U-net prediction accuracy in each chamber group to further improve the prediction accuracy. In contrast to the earlier state-space model proposed for cardiac motion estimation \citep{punithakumar2010regional}, the proposed method utilizes an angular time-varying frequency component. The input to the filter is set as contour points in Cartesian coordinates obtained from predicted image masks after boundary detection and uniform sampling. The UKF first iterates over the uniformly sampled points and then iterates over entire frames in the cardiac cycle. The overall iterative process detailing the UKF steps for each contour point is shown in Fig.~\ref{fig:ukf_mainfig}. The UKF relies on a state transition model to enforce temporal smoothing. The UKF utilizes the Kalman gain, which is computed at each iteration in the process to deal with measurement uncertainty resulting from the U-net predictions.

\subsubsection{Uniform sampling of contour points}
The semantic segmentation results for the 2, 3 and 4-chamber sequences by U-net models were converted into contour points. Each contour corresponding to different cardiac frames is sampled to 60 points, and the sampling is performed by transforming the contours to the polar coordinate system. The origin of the polar coordinate system is selected as the centroid of the segmented region and the sampling points are selected at $6^\circ$ intervals along the theta-axis using linear interpolation. Centroid $(x_c,y_c)$ of the segmented region is computed using \eqref{eq32}, where $n$ is the number of raw contour points and $(x_p,y_p)$ is the $p^{\textrm{th}}$ point on raw contour set

\begin{equation}
    x_c=\sum_{p=1}^n x_p\,, \label{eq32} \qquad y_c=\sum_{p=1}^n y_p\,.
\end{equation}

We convert each point $(x_p, y_p)$ on the automated left atrial contour from the Cartesian coordinate system to the polar coordinate system using the following equations:

\begin{equation}
    r_p = \displaystyle\sqrt{x_p^2+y_p^2}\label{eq31}\,, \qquad \theta_p = \tan^{-1}\left(\frac{y_p}{x_p}\right)\,.
\end{equation}

A linear interpolation is then applied to find uniformly sampled points along the angular coordinates in the polar coordinate system. These points in the polar coordinate system are converted back to the Cartesian coordinate system and used in the subsequent processing using unscented Kalman filter.

\subsection{Dynamic model for temporal periodicity}
A dynamic model is designed to characterize cardiac motion. Assume that $(x,y)$ be one of the points on contour obtained after resizing the prediction masks of U-net to original image size and consider the corresponding state vector $\zeta=[\bar{x} \; x \; \dot{x}]$ which describes the dynamics of point in x-coordinate direction. The terms $\bar{x}$ and $\dot{x}$ denote mean position and velocity over the entire cardiac cycle, respectively. We exploit the periodicity feature in each cardiac cycle and propose a state-space model \eqref{eq6} corresponding to the oscillation:
\renewcommand{\arraystretch}{1.5}
\begin{equation}
\dot{\zeta}(t)=
\begin{bmatrix}
0 & 0 & 0 \\
0 & 0 & 1 \\
\omega^2 & -\omega^2 & 0\\
\end{bmatrix}
\zeta(t)
+
\begin{bmatrix}
1 & 0 \\
0 & 0 \\
0 & 1 \\
\end{bmatrix}
w(t)\,,
\label{eq6}
\end{equation}
\renewcommand{\arraystretch}{1.0}
where $\omega$ is the angular frequency and $w(t)$ is the vector-valued white noise. The model describes the cardiac motion as a simple harmonic motion. Nonetheless, cardiac movement is much more complex than a simple harmonic motion and a time-varying angular velocity is added to account for the rate of oscillation variation and the discrete equivalence of the model is described in 

\begin{equation}
    \zeta_{k+1}=F_k\zeta_k+w_k\,,
    \label{eq7}
\end{equation}
where $k$ is the frame number between each heart beat and the maximum value $K$ denotes the number of frames in a cardiac cycle which is typically between 25 to 30. $F_k$ is given in the following: 
\begin{equation}
F_k=
\renewcommand{\arraystretch}{1.5}
\small{\begin{bmatrix}
1 & 0 & 0\\
1-\cos(\omega_k\Delta T)&\cos(\omega_k\Delta T) & \frac{1}{\omega_k}\sin(\omega_k\Delta T)\\
\omega_k\sin(\omega_k\Delta T) & -\omega_k\sin(\omega_k\Delta T) & \cos(\omega_k\Delta T)
\end{bmatrix}}\,,
\label{eq8}
\end{equation}
where $w_k$ is a discrete-time white noise, $\Delta T$ is the sampling interval and $\omega_k$ is the discrete angular frequency.
The initial estimated value of the angular frequency $\omega_1$ is computed as
\begin{equation}
\omega_1=\frac{2\pi \times \textrm{Heart Rate}}{60}\,,
\label{eq9}
\end{equation}
and the time interval between two consecutive cardiac frames $\Delta T$ is calculated as
\begin{equation}
\Delta T= \frac{60}{\textrm{Heart Rate}\times K}\,.
\label{eq10}
\end{equation}
The covariance of process noise $Q_k$ is given by
\begin{equation}
Q_k=[q_{ij}]_{3\times3}\,,
\label{eq11}
\end{equation}
where $q_{ij}$ is defined in \eqref{eq12}--\eqref{eq17}:
\renewcommand{\arraystretch}{1.5}
\begin{align}
	q_{11} & = q^2_1 \Delta T \label{eq12}\,,\\
    q_{12} & = q_{21} = {\frac {q^2_1 ( \omega_k \Delta T-\sin (\omega_k \Delta T))}{\omega_k}}\label{eq13}\,,\\
    q_{13} & = q_{31} = q^2_1 (1-\cos ( \omega_k \Delta T ))\label{eq14}\,,\\
    q_{22} & = {\frac{
		\begin{array}{l}
			q^2_1\omega_k^2(3\omega_k \Delta T-4\sin(\omega_k \Delta T)\\
            \qquad+\,\cos(\omega_k \Delta T)\sin(\omega_k \Delta T))\\
			\qquad\qquad +\,q^2_2 (\omega_k \Delta T - \cos(\omega_k \Delta T)\sin(\omega_k \Delta T))
		\end{array}}{2\omega_k^3}}\label{eq15}\,,\\
        \renewcommand{\arraystretch}{1.5}
    q_{23} & = q_{32} = {\frac{
		\begin{array}{l}
			q^2_1\omega_k^2(1-2\cos(\omega_k \Delta T) +\cos^2(\omega_k \Delta T))\\
			\qquad +\,q^2_2\sin^2(\omega_k \Delta T)
		\end{array}}{2\omega_k^2}}\,,\label{eq16}\\	 
	q_{33} & = {-\frac{
		\begin{array}{l}
				q^2_1\omega_k^2(\cos(\omega_k \Delta T)\sin(\omega_k \Delta T) -\omega_k \Delta T)\\
				\qquad -\,q^2_2(\cos(\omega_k \Delta T)\sin( \omega_k \Delta T) +\omega_k \Delta T)
		\end{array}}{2\omega_k}} \,,\label{eq17} 
\end{align}
\renewcommand{\arraystretch}{1.0}
where $q_1$ and $q_2$ describe uncertainties associated with the mean position and velocity respectively. Small $q_1$ and $q_2$ will enforce the conformity of cardiac motion to the model, thereby affect the accuracy of the prediction results when motion deviates from the model. Large values of $q_1$ and $q_2$ allow for high uncertainties within the model, which may lead to poor temporal consistency. In our case, the noise parameters $q_1=q_2=1\times10^{-3}$ are chosen empirically.

Let $S_k=[\bar{x}_k \;x_k \; \dot{x}_k \; \bar{y}_k \; y_k \; \dot{y}_k \; \omega_k]^T$ be the state vector that describes the corresponding dynamics at time frame $k$. Elements $\bar{x}$, $\dot{x}$, $\bar{y}$, $\dot{y}$ denote, respectively, the mean position and velocity in x-coordinate and mean position and velocity in y-coordinate. The discrete-time model that describes the cyclic motion of the point is given in 
\begin{equation}
S_{k+1}=f_k(S_k)+v_k\,,
\label{eq18}
\end{equation}
where $f_k(S_k)$ is defined by
\begin{equation}
f_k(S_k)=
\renewcommand{\arraystretch}{1.5}
\begin{bmatrix}
F_k & 0_{3\times 3} & 0_{3\times 1}\\
0_{3\times 3} & F_k & 0_{3\times 1}\\
0_{3\times 3} & 0_{3\times 3} & 1
\end{bmatrix}S_k\,,
\renewcommand{\arraystretch}{1.0}
\label{eq19}
\end{equation}
and $v_k$ denotes a Gaussian process noise sequence with zero-mean and covariance that accommodates unpredictable error due to model uncertainties given in 
\begin{equation}
\text{cov}(v_k)=
\renewcommand{\arraystretch}{1.5}
\begin{bmatrix}
Q_k & 0_{3\times 3} & 0_{3\times 1}\\
0_{3\times 3} & Q_k & 0_{3\times 1}\\
0_{3\times 3} & 0_{3\times 3} & 1
\end{bmatrix}
\renewcommand{\arraystretch}{1.0}\,,
\label{eq20}
\end{equation}
where $Q_k$ is defined in \eqref{eq11}. The measurement equation is given in 
\begin{equation}
\mathcal{Z} = H_kS_k + \eta_k\,,
\label{eq21}
\end{equation}
where $H_k$ is defined in 
\begin{equation}
H_k=
\renewcommand{\arraystretch}{1.2}
\begin{bmatrix}
0 & 1 & 0 & 0 & 0 & 0 & 0\\
0 & 0 & 0 & 0 & 1 & 0 & 0
\end{bmatrix}
\renewcommand{\arraystretch}{1.0}\,,
\label{eq22}
\end{equation}
and $\eta_k$ is a zero-mean Gaussian noise sequence to model the measurement uncertainties such as potential false positives resulting from inaccurate U-net predictions, with covariance demonstrated in 
\begin{equation}
\text{cov}(\eta_k)=
\renewcommand{\arraystretch}{1.2}
\begin{bmatrix}
r & 0 \\
0 & r
\end{bmatrix}
\renewcommand{\arraystretch}{1.0}\,.
\label{eq23}
\end{equation}
The parameter $r$ characterizes the uncertainties associated with observations. Smaller values of $r$ enforce the conformity of the estimation to the measurements while larger values allow for associating more uncertainty with the measurements. In the proposed study, U-net based model showed adequate raw accuracy of the prediction masks and $r$ was set to $1\times10^{-2}$. Apart from the measurement uncertainties modeled using $\eta_k$, the UKF updates its state and covariance based on the relative ratio of the belief in measurement over the belief in state by calculating Kalman gain shown in Fig.~\ref{fig:ukf_mainfig} to improve prediction performance. If the Kalman gain is high, the filter will rely more on the current measurement. Otherwise, it will rely more on the predicted value by the filter.

\subsection{Filter initialization}
It is not possible to have prior information of the exact initial value of $S_1$ in the proposed study. Therefore, a two-point differencing method proposed by \citet{bar2004estimation} is applied to initialize the position and velocity components of the state. For instance, the initial position $\hat{x_1}$ and velocity $\hat{\dot{x_1}}$ in x-coordinate of the $i$th sample point: 
\begin{equation}
\hat{x_1}=z_1^{ix}\,,
\label{eq24}
\end{equation}
\begin{equation}
\hat{\dot{x_1}}=\displaystyle\frac{z_2^{ix}-z_1^{ix}}{\Delta T}\,,
\label{eq25}
\end{equation}
where $z_k^{ix}, k=1,2,...,K$ is the observation in x-coordinate obtained from $k$th frame. The mean position over cardiac cycle illustrated in \eqref{eq26} was initialized by taking the average of observation in x-coordinate
\begin{equation}
\hat{\bar{x_1}}=\frac{1}{K}\sum_{k=1}^{K} z_k^{ix}\,.
\label{eq26}
\end{equation}
Similarly, the initial state elements in y-coordinate can be computed using the same strategy and the final initial state vector input $S_1$ is given in \eqref{eq27}
\begin{equation}
S_1 = [\hat{\dot{x_1}} \; \hat{x_1} \; \hat{\bar{x_1}} \; \hat{\dot{y_1}} \; \hat{y_1} \; \hat{\bar{y_1}} \; \omega_1]^T\,.
\label{eq27}
\end{equation}
The corresponding initial covariance is given in 
\begin{equation}
P_1=
\renewcommand{\arraystretch}{1.5}
\begin{bmatrix}
\phi_1 & 0_{3\times 3} & 0_{3\times 1}\\
0_{3\times 3} & \phi_1 & 0_{3\times 1}\\
0_{3\times 3} & 0_{3\times 3} & 1
\end{bmatrix}
\renewcommand{\arraystretch}{1.0}\,,
\label{eq28}
\end{equation}
where $\phi_1$ is given in 
\begin{equation}
\phi_1 = 
\renewcommand{\arraystretch}{1.5}
\begin{bmatrix}
r & \frac{r}{k} & \frac{r}{k \Delta T}\\
\frac{r}{k} & r & \frac{r}{\Delta T}\\
\frac{r}{k \Delta T} & \frac{r}{\Delta T} & \frac{2r}{\Delta T^2}
\end{bmatrix}
\renewcommand{\arraystretch}{1.0}\,.
\label{eq29}
\end{equation}

\subsection{Iterative prediction and update}
After generating sigma points using the method proposed in \citep{Julier2004}, prediction and update steps were performed iteratively based on \eqref{eq18} and \eqref{eq21}. The entire process of the UKF is summarized in Algorithm~\ref{alg_ukf}.

\section{Experimental Setup and Results}

\subsection{Dataset, evaluation criteria and implementation details}
The U-net and the proposed algorithms were trained and tested over 100 MRI patient datasets acquired retrospectively at the University Alberta Hospital. The study was approved by the Health Research Ethics Board of the University of Alberta. The details of the datasets used in the proposed model are given in Table \ref{tab1}. The evaluations were performed over 1515 images acquired from 20 patients in comparison to the expert manual segmentation of the left atrium from the 2, 3 and 4-chamber cine steady state processions long-axis MR sequences where each chamber group consists of 505 images. The 2, 3 and 4-chamber sequences are shortened as \textit{2ch}, \textit{3ch} and \textit{4ch} in the rest of the paper. The ground truth manual segmentation was initially performed by a medical student using a semi-automated software and the contours were corrected by an experienced radiologist.

\begin{table*}[htbp]
\setlength{\tabcolsep}{10pt}
\caption{Details of data set used in the proposed study.}
\renewcommand{\arraystretch}{1.5}
\centering
\begin{tabular}{lcc}
\toprule
\textbf{Description}                   & \textbf{Training Set}              & \textbf{Test Set}             \\ \hline
\textbf{Number of subjects}            & $20-80$                            &  20                           \\ \hline
\textbf{Patient's sex}                 & 40 Males / 40 Females              & 10 Males / 10 Females         \\ \hline
\textbf{Scanner protocol}              & Siemens AERA                       & Siemens AERA                  \\ \hline
\textbf{Patient's age}                 & $4-85$ years                       & $29-73$ years                 \\ \hline
\textbf{Magnetic strength}             & 1.5 T                              & 1.5 T                         \\ \hline
\textbf{Number of frames ($K$)} & 25 or 30                           & 25 or 30                      \\ \hline
\textbf{Image size}                    & (176, 184) $-$ (256, 208) pixels   & (192, 208) $-$ (256, 256) pixels \\ \hline
\textbf{Pixel spacing}                 & $1.445-1.797$ mm                   & $1.445-1.660$ mm              \\ \hline
\textbf{Repetition time (TR)}          & $21.12-89.10$ ms                   & $27.10-67.25$ ms              \\ \hline
\textbf{Echo time (TE)}                & $1.09-1.43$ ms                     & $1.12-1.18$ ms                \\ \bottomrule
\end{tabular}
\label{tab1}
\end{table*}

In order to empirically evaluate the impact of training sample size on the algorithm performance, four different scenarios were created by increasing the sample size from 20 to 80 patients with an increment of 20 patients per scenario (Refer to Fig.~\ref{fig1}). These scenarios are shortened as \textit{Train 20-80} in the rest of the paper. For each case, the U-net based neural nets were trained and tested using 5-fold of cross validation to prevent over-fitting. The training loss function was chosen as the negative of the Dice coefficient. To demonstrate the effectiveness of the proposed approach, the results were compared with the algorithm that was trained using a merged set of all cardiac long-axis chambers. 

\subsection{Implementation}
The U-net based models, \textit{Classnet-A} and \textit{Classnet-B} were implemented in Python programming language using Keras machine learning module with Tensorflow backend and cudnn. The filterpy Python module \citet{labbe2014kalman} was used for the implementation of Bayesian based unscented Kalman filter. Three U-net based architectures were utilized with linear variable learning rate with Adam optimizer from $1\times10^{-4}$ in initial to $1\times10^{-5}$ for \emph{2ch} U-net and $1\times10^{-5}$ in initial to $1\times10^{-6}$ for \emph{3ch} and \emph{4ch}. The decay was uniform per each epoch. Batch size was set as 32 and epoch number is set to 500 in a 5-fold cross validation scheme. The proposed model was tested on a desktop with Intel core i7-7700 HQ CPU with 16 GB RAM. The neural net models were trained and tested with an NVIDIA GTX 1080 Ti graphics processing unit with 12 GB memory. The approximate training time for each chamber group is 7 hours and the prediction time is 4 ms per image for the proposed method.

\subsection{Quantitative Evaluation Metrics}
The proposed method and the U-net based approach were evaluated quantitatively using Dice coefficient (DC), Hausdorff distance (HD) and reliability metric.

\subsubsection{Dice coefficient}
Dice coefficient illustrated in \eqref{eq1} measures the overlap between two delineated regions, where set $A$ is the prediction image set and $M$ is the manually labelled image set
\begin{equation}
    DC=\frac{2\vert A\cap M\vert}{\vert A \vert + \vert M \vert}\,.
    \label{eq1}
\end{equation}

\subsubsection{Hausdorff distance}
The Hausdorff distance measures the maximum deviation between two contours in terms of the Euclidean distance. The Hausdorff distance between manual and automatic contours was computed as follows:
\begin{multline}
    d_H(a,m) = \displaystyle\max\bigg\{\max_i\left(\min_j\left(d(p^i_a,p^j_m) \right) \right),\\
    \displaystyle\max_j\left(\min_i\left(d(p^i_a,p^j_m)\right)\right) \bigg\}\,,
\end{multline}
where $d(\cdot)$ is the Euclidean distance, $\{p^i_a\}$ denotes the set of  points on automated contour and $\{p^j_m\}$ denotes the set of points on manual contour. 

\subsubsection{Reliability metric}
The reliability metric is evaluated using the function given in \eqref{eq4}. The complementary cumulative distribution function is defined for each $d\in[0,1]$ as the probability of obtaining Dice coefficient higher than $d$ over all images
\begin{equation}
R(d)=P_r(DC>d)=\frac{\begin{array}{l}
     \textrm{\small Images with Dice}  \\
     \quad \textrm{\small coefficient higher than $d$}
\end{array} }{\textrm{\small Total images}}\,.
\label{eq4}
\end{equation}

\subsection{Experimental results}

\begin{figure*}[htbp]
    \centering
    \begin{tabular}{c c c}
         \includegraphics[scale=0.62]{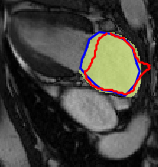} &
         \includegraphics[scale=0.74]{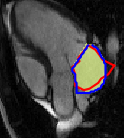} &
         \includegraphics[scale=0.51]{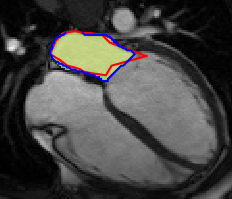}
         \vspace{6pt}\\
         {\small 2ch Train 80} & {\small 3ch Train 80} & {\small 4ch Train 80}\\
         {\small DC: U-net+UKF $=93.3\%$} & {\small DC: U-net+UKF $=94.1\%$} & {\small DC: U-net+UKF $=95.2\%$}\\
         {\small DC: U-net $=91.7\%$} & {\small DC: U-net $=89.6\%$} & {\small DC: U-net $=91.5\%$}\\
    \end{tabular}
    \caption{Representative example results by the proposed method (blue curve) and U-net (red curve) approaches in comparison to the expert manual ground truth (yellow area). The proposed method yielded more conformity with manual ground truth than U-net approach. }
    \label{fig13}
\end{figure*}

\begin{table*}[]
\caption{Classification result for \textit{Classnet-A} proposed in this paper. The test images consisted of 505 images for each chamber group. Overall accuracy is reported for each training scenario in the last row.}
\centering
\renewcommand{\arraystretch}{1.2}
\begin{tabular}{c c c c| c c c| c c c| c c c}
\toprule
\multicolumn{1}{l}{\multirow{3}{*}{\textbf{Ground Truth}}} & \multicolumn{12}{c}{\textbf{Prediction Result}}                                                                                                                                                                                                                                                                                  \\ \cline{2-13} 
\multicolumn{1}{l}{}                                       & \multicolumn{3}{c|}{\textbf{Train 20}}                                         & \multicolumn{3}{c|}{\textbf{Train 40}}                                         & \multicolumn{3}{c|}{\textbf{Train 60}}                                         & \multicolumn{3}{c}{\textbf{Train 80}}                                         \\ \cline{2-13} 
\multicolumn{1}{l}{}                                       & \multicolumn{1}{l}{2ch} & \multicolumn{1}{l}{3ch} & \multicolumn{1}{l|}{4ch} & \multicolumn{1}{l}{2ch} & \multicolumn{1}{l}{3ch} & \multicolumn{1}{l|}{4ch} & \multicolumn{1}{l}{2ch} & \multicolumn{1}{l}{3ch} & \multicolumn{1}{l|}{4ch} & \multicolumn{1}{l}{2ch} & \multicolumn{1}{l}{3ch} & \multicolumn{1}{l}{4ch} \\ \hline
2ch                                                          & \textbf{505}             & 0                       & 0                        & \textbf{505}             & 0                       & 0                        & \textbf{505}             & 0                       & 0                        & \textbf{505}             & 0                       & 0                        \\ \hline
3ch                                                          & 20                        & \textbf{485}             & 0                        & 0                        & \textbf{480}             & 25                        & 0                        & \textbf{505}             & 0                       & 0                        & \textbf{505}             & 0                        \\ \hline
4ch                                                          & 0                        & 24                       & \textbf{481}             & 0                        & 17                       & \textbf{488}             & 0                        & 30                       & \textbf{475}             & 0                        & 0                        & \textbf{505}             \\ \hline
\textit{\textbf{Accuracy(\%)}}                               & \multicolumn{3}{c}{97.1}                                                      & \multicolumn{3}{c}{97.2}                                                      & \multicolumn{3}{c}{98.0}                                                      & \multicolumn{3}{c}{\textbf{100.0}}                                             \\ \bottomrule

\end{tabular}
\label{tab7}
\end{table*}

\begin{table*}[]
\caption{Classification result for \textit{Classnet-B} proposed in this paper. The test images consisted of 1515 images from all chamber groups (each consists of 505 images). Class 1 indicates the images that need to be post-processed using the UKF and Class 2 indicates the images that do not need to be post-processed due to their higher accuracy. Overall accuracy is reported for each training scenario in the last row.}
\centering
\renewcommand{\arraystretch}{1.2}
\begin{tabular}{c c c | c c | c c | c c}
\toprule
\multicolumn{1}{l}{\multirow{3}{*}{\textbf{Ground Truth}}} & \multicolumn{8}{c}{\textbf{Prediction Result}}                                                                                                                                                                  \\ \cline{2-9} 
\multicolumn{1}{l}{}                               & \multicolumn{2}{c|}{\textbf{Train 20}}              & \multicolumn{2}{c|}{\textbf{Train 40}}              & \multicolumn{2}{c|}{\textbf{Train 60}}              & \multicolumn{2}{c}{\textbf{Train 80}}                 \\ \cline{2-9} 
\multicolumn{1}{l}{}                               & \multicolumn{1}{l}{Class 1} & \multicolumn{1}{l|}{Class 2}  & \multicolumn{1}{l}{Class 1} & \multicolumn{1}{l|}{Class 2}  & \multicolumn{1}{l}{Class 1} & \multicolumn{1}{l|}{Class 2}  & \multicolumn{1}{l}{Class 1} & \multicolumn{1}{l}{Class 2}    \\ \hline
Class 1                                            & \textbf{1167}            & 3                         & \textbf{634}            & 0                         & \textbf{542}            & 0                         & \textbf{338}            & 0               \\ \hline
Class 2                                            & 52                     & \textbf{293}              & 0                       & \textbf{881}              & 0                       & \textbf{973}              & 0                       & \textbf{1177}          \\\hline 

\textit{\textbf{Accuracy(\%)}}                               & \multicolumn{2}{c}{96.4}                                                      & \multicolumn{2}{c}{100.0}                                                      & \multicolumn{2}{c}{100.0}                                                      & \multicolumn{2}{c}{\textbf{100.0}}                                             \\ \bottomrule
\end{tabular}
\label{tab:classnetb}
\end{table*}

\subsubsection{Quantitative assessment}
The \textit{Classnet-A} and \textit{Classnet-B}'s performance are reported in Table~\ref{tab7} and Table~\ref{tab:classnetb}, respectively. Both classification networks achieved $100\%$ accuracy in the \textit{Train 80} scenario where the approaches have correctly classified all 1515 images to the respective chamber group or image classification into Class 1 and Class 2 based on U-net segmentation accuracy. The results from both tables show that the accuracy of the classification algorithms improves with the size of training set. The number of classes on ground truth remained the same for Train 20, 40, 60, and 80 scenarios for \textit{Classnet-A}, as indicated in Table~\ref{tab7}, since it is based on the chamber views on the test set. However, the number of classes on ground truth varied for different scenarios for \textit{Classnet-B}, as indicated in Table~\ref{tab:classnetb}, due to varying U-net prediction accuracy on the testing set. For instance, the U-net trained with Train 80 scenario resulted in more images ($n=1177$) in Class 2 (Dice score $\geq 0.9$) than the one trained with Train 20 ($n=345$). 

The quantitative assessment of the automated segmentation results for \emph{2ch}, \emph{3ch} and \emph{4ch} are given in Tables \ref{tab8}, \ref{tab9}, and \ref{tab10}, respectively. The tables report the mean and standard deviation of Dice coefficient and Hausdorff distance values for U-net+UKF and U-net alone approaches with and without the application of \emph{Classnet-A} neural net for different sizes of training data sets. As shown in Tables \ref{tab8}--\ref{tab10}, the proposed U-net+UKF method with \emph{Classnet-A} outperformed U-net for all three chamber groups in terms of Hausdorff distance in \textit{Train 80} scenario. The method also outperformed U-net in \emph{2ch} and \emph{3ch} sequences in \textit{Train 80} scenario. The reported values in Tables \ref{tab8}--\ref{tab10} demonstrate that the proposed U-net+UKF approach outperformed U-net alone method in terms of Dice coefficient in 23 out of 24 different tested scenarios that consist of with and without \emph{Classnet-A}, \emph{Train 20} -- \emph{Train 80} scenarios, and \emph{2ch} -- \emph{4ch} chamber groups. The reported values also demonstrate that the \emph{Classnet-A} approach improved performance in terms of Dice coefficient in  8 out of 12 scenarios tested with different chamber groups and training sets. 

The best performances for each chamber group in terms of Dice coefficient and Hausdorff distance were obtained using the proposed U-net+UKF using \emph{Classnet-A} approach with \emph{Train 80} except for the \emph{4ch} segmentation measured in terms of Dice score where a small decrease of 0.1\% in performance was observed with the application of \emph{Classnet-A}. The approach yielded average Dice coefficients of 94.1\%, 93.7\% and 90.1\% for \emph{2ch}, \emph{3ch}, and \emph{4ch}, respectively. The corresponding average Hausdorff distance values were 6.0 mm, 5.5 mm and 8.4 mm, respectively.

The accuracy of the proposed U-net+UKF with \emph{Classnet-A} method steadily increased with the size of the training data for all chamber groups as reported in Tables \ref{tab8}--\ref{tab10}. The increase in accuracy with the size of training data set is also observed in the reliability assessment as depicted in Fig.~\ref{fig11}. The improvement was significant from \textit{Train 20} to \textit{Train 40} in 2-chamber and marginal from \textit{Train 40} to \textit{Train 80}. In 3-chamber group, the improvement was critical from \textit{Train 40} to \textit{Train 60} while in 4-chamber group the overall improvement is marginal with the training size.

\subsubsection{Visual assessment}
\begin{figure*}[htbp]
    \centering
    \begin{tabular}{c c c c c}
        \includegraphics[width=0.373\columnwidth, trim=2.5cm 2.0cm 2.5cm 2.5cm,clip]{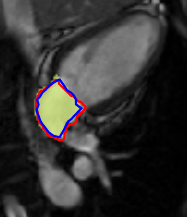} &
        \includegraphics[width=0.373\columnwidth, trim=2.5cm 2.0cm 2.5cm 2.5cm,clip]{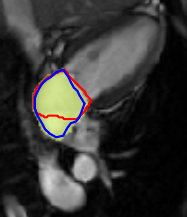} & \includegraphics[width=0.373\columnwidth, trim=2.5cm 2.0cm 2.5cm 2.5cm,clip]{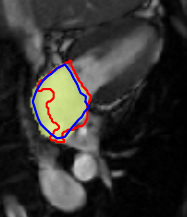} & \includegraphics[width=0.373\columnwidth, trim=2.5cm 2.0cm 2.5cm 2.5cm,clip]{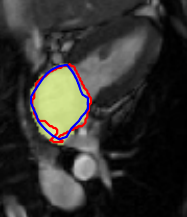} & \includegraphics[width=0.373\columnwidth, trim=2.5cm 2.0cm 2.5cm 2.5cm,clip]{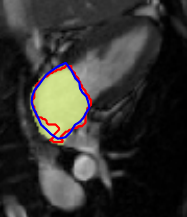}  \vspace{6pt}\\
        
        {\small 2ch Frame 1} & {\small 2ch Frame 6} & {\small 2ch Frame 11} & {\small 2ch Frame 16} & {\small 2ch Frame 21} \\
        {\small Proposed$=96.0\%$} & {\small Proposed $=96.2\%$} & {\small Proposed $=96.0\%$} & {\small Proposed $=95.1\%$} & {\small Proposed $=94.1\%$} \\
        {\small U-net $=94.2\%$} & {\small U-net $=83.7\%$} & {\small U-net $=82.4\%$} & {\small U-net $=91.7\%$} & {\small U-net $=91.6\%$} \vspace{6pt}\\
        
        \includegraphics[width=0.373\columnwidth, trim=2.0cm 2.6cm 3.8cm 2.8cm,clip]{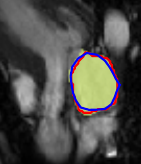} & \includegraphics[width=0.373\columnwidth, trim=2.0cm 2.6cm 3.8cm 2.8cm,clip]{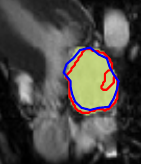} & \includegraphics[width=0.373\columnwidth, trim=2.0cm 2.6cm 3.8cm 2.8cm,clip]{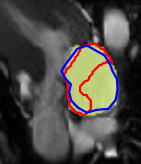} & \includegraphics[width=0.373\columnwidth, trim=2.0cm 2.6cm 3.8cm 2.8cm,clip]{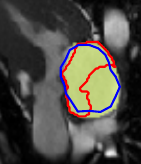} & \includegraphics[width=0.373\columnwidth, trim=2.0cm 2.6cm 3.8cm 2.8cm,clip]{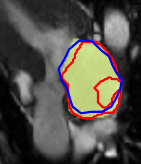} \vspace{6pt}\\
        
        {\small 3ch Frame 1} & {\small 3ch Frame 6} & {\small 3ch Frame 11} & {\small 3ch Frame 16} & {\small 3ch Frame 21}\\
        {\small Proposed$=93.2\%$} & {\small Proposed $=93.5\%$} & {\small Proposed $=93.0\%$} & {\small Proposed $=91.3\%$} & {\small Proposed $=92.5\%$} \\
        {\small U-net $=95.7\%$} & {\small U-net $=92.7\%$} & {\small U-net $=68.8\%$} & {\small U-net $=71.4\%$} & {\small U-net $=88.9\%$} \vspace{6pt}\\
        
        \includegraphics[width=0.373\columnwidth, trim=1.8cm 3.3cm 2.8cm 1.2cm,clip]{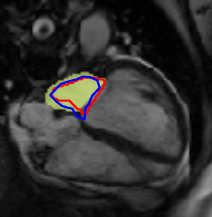} & \includegraphics[width=0.373\columnwidth, trim=1.8cm 3.3cm 2.8cm 1.2cm,clip]{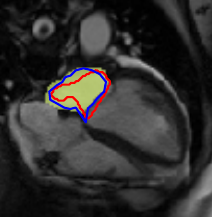} & \includegraphics[width=0.373\columnwidth, trim=1.8cm 3.3cm 2.8cm 1.2cm,clip]{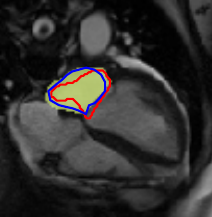} & \includegraphics[width=0.373\columnwidth, trim=1.8cm 3.3cm 2.8cm 1.2cm,clip]{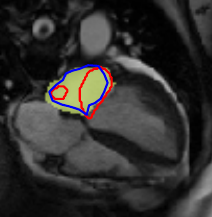} & \includegraphics[width=0.373\columnwidth, trim=1.8cm 3.3cm 2.8cm 1.2cm,clip]{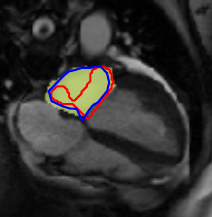}  \vspace{6pt}\\
        
        {\small 4ch Frame 1} & {\small 4ch Frame 6} & {\small 4ch Frame 11} & {\small 4ch Frame 16} & {\small 4ch Frame 21} \\
        {\small Proposed$=86.8\%$} & {\small Proposed $=90.7\%$} & {\small Proposed $=90.5\%$} & {\small Proposed $=91.2\%$} & {\small Proposed $=91.3\%$} \\
        {\small U-net $=76.9\%$} & {\small U-net $=71.9\%$} & {\small U-net $=73.5\%$} & {\small U-net $=66.0\%$} & {\small U-net $=72.4\%$} \vspace{6pt}\\
        
    \end{tabular}
    \caption{An example showing the prediction results on selected frames over a complete cardiac cycle (K=25) for 2, 3 and 4-chamber views (from top to bottom) where the U-net yielded temporally inconsistent segmentation results. The yellow area is the manually labelled left atrium. The blue and red curves are the prediction results by the proposed U-net+UKF and U-net alone, respectively. A higher conformance was observed between expert manual segmentation and the results by the proposed fully automated approach which demonstrates the benefits of applying temporal smoothing with the UKF.}
    \label{fig_unet_ukf_cmp}
\end{figure*}

\begin{figure}[htbp]
    \centering
    \begin{tabular}{c c}
         \includegraphics[trim=1.5cm 2.5cm 3.2cm 2.3cm,clip]{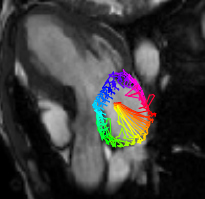} 
         &  \includegraphics[trim=1.5cm 2.5cm 3.2cm 2.3cm,clip]{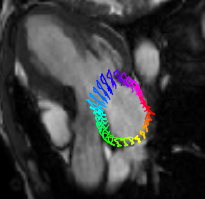}\vspace{6pt}\\
         
        {\small (a) U-net} & {\small (b) U-net+UKF} \vspace{6pt}
        
    \end{tabular}
    \caption{An example showing the trajectories of the left atrial boundary points over the entire cardiac cycle (K=25) estimated using U-net alone, and U-net+UKF approaches. The example demonstrates the benefits of applying the UKF to enforce the temporal consistency in estimating the contour points as the method allows for correcting inaccurate segmentation results by the U-net in a few arbitrary image frames.}
    \label{fig:contour_sample_frame}
\end{figure}

\begin{figure*}[htbp]
    \centering
    \begin{tabular}{c c c c c}
        \includegraphics[scale=0.48]{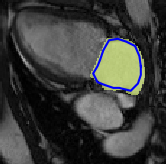} &
        \includegraphics[scale=0.48]{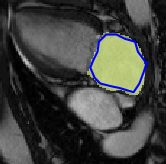} & \includegraphics[scale=0.48]{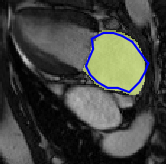} & \includegraphics[scale=0.48]{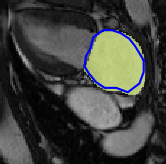} & \includegraphics[scale=0.48]{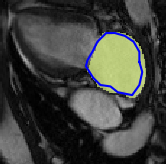}  \vspace{6pt}\\
        {\small 2ch Frame 1} & {\small 2ch Frame 6} & {\small 2ch Frame 11} & {\small 2ch Frame 16} & {\small 2ch Frame 21} \vspace{6pt}\\
        \includegraphics[scale=0.65]{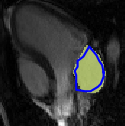} & \includegraphics[scale=0.65]{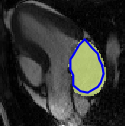} & \includegraphics[scale=0.65]{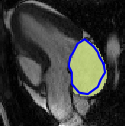} & \includegraphics[scale=0.65]{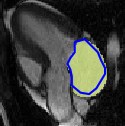} & \includegraphics[scale=0.65]{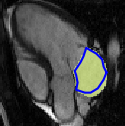}  \vspace{6pt}\\
        {\small 3ch Frame 1} & {\small 3ch Frame 6} & {\small 3ch Frame 11} & {\small 3ch Frame 16} & {\small 3ch Frame 21}\vspace{6pt}\\
        \includegraphics[scale=0.39]{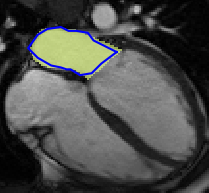} & \includegraphics[scale=0.39]{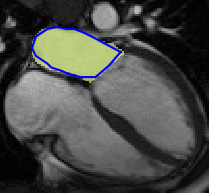} & \includegraphics[scale=0.39]{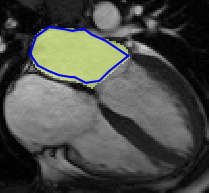} & \includegraphics[scale=0.39]{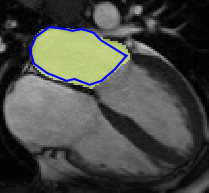} & \includegraphics[scale=0.39]{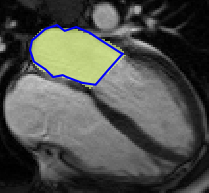}  \vspace{6pt}\\
        {\small 4ch Frame 1} & {\small 4ch Frame 6} & {\small 4ch Frame 11} & {\small 4ch Frame 16} & {\small 4ch Frame 21}
    \end{tabular}
    \caption{An example showing the prediction results over a complete cardiac cycle (25 frames in total) for 2, 3 and 4-chamber groups (from top to bottom). The yellow area is the manually labelled region of interest corresponds to left atrium and the blue curve is the prediction results by the proposed U-net+UKF. A higher conformance was observed between expert manual segmentation and the results by the proposed fully automated approach}
    \label{fig8}
\end{figure*}

\begin{figure*}[htbp]
    \centering
    \begin{tabular}{ccc}
        \includegraphics[scale=0.32]{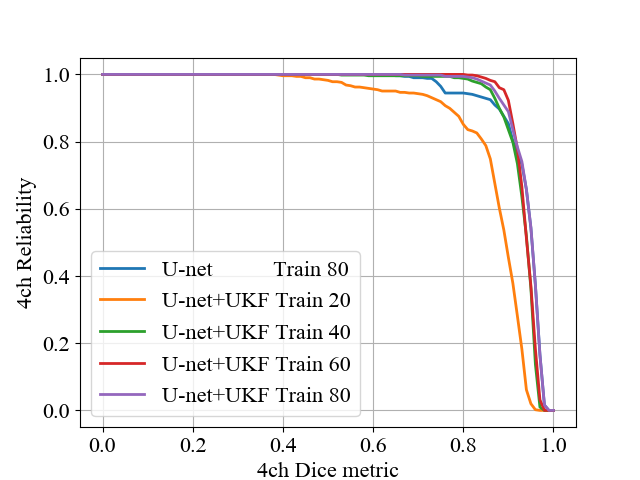} & \includegraphics[scale=0.32]{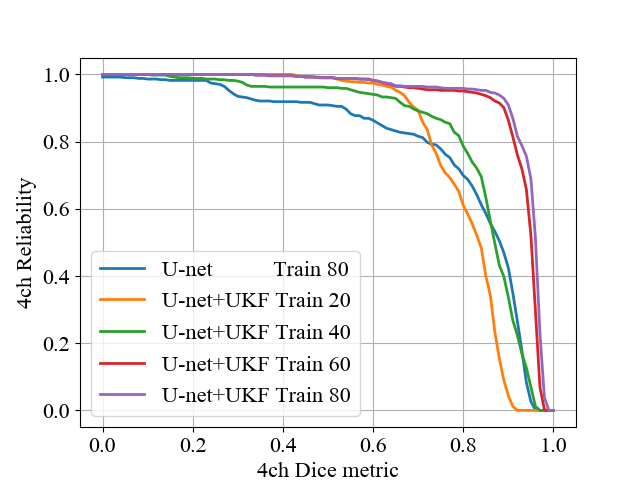} & \includegraphics[scale=0.32]{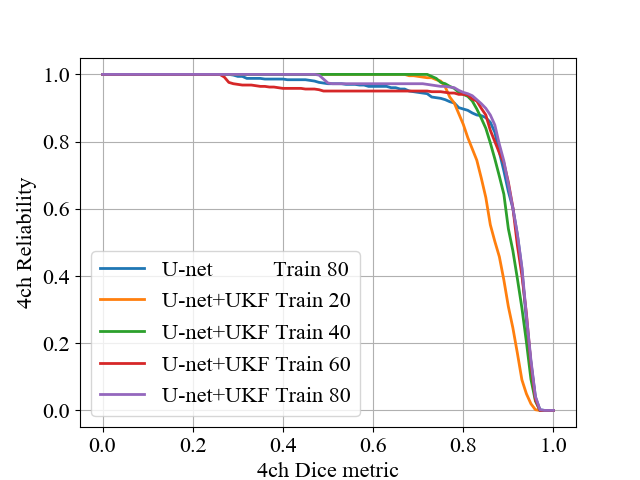}  \\
        {\small 2-chamber reliability} & {\small 3-chamber reliability} & {\small 4-chamber reliability}
    \end{tabular}
    \caption{Reliability versus Dice metric for the proposed U-net+UKF with \emph{Classnet-A} approach with varying training set data. The training were performed using data sets acquired from 20, 40, 60 and 80 patients for each chamber group. The evaluations were performed over 505 images for each chamber group acquired from 20 patients.}
    \label{fig11}
\end{figure*}

Fig.~\ref{fig13} shows example automatic contours by the proposed U-net+UKF approach (blue curve) as well as U-net (red curve) with the application of \emph{Classnet-A}. The manual ground truth delineation is depicted by the yellow area. Fig.~\ref{fig_unet_ukf_cmp} shows comparative examples of U-net and U-net+UKF prediction results on selected frames over a complete cardiac cycle for 2, 3, and 4 chamber views where the U-net yielded temporally inconsistent results. The examples show that the proposed application of temporal smoothing for image sequences consisting of image frames with inaccurate U-net predictions led to improved segmentation accuracy. Fig.~\ref{fig:contour_sample_frame} shows the trajectories of left atrial boundary points over the complete cardiac cycle, which demonstrates the advantage of applying UKF for better temporal consistency compared to U-net alone. As shown in Fig.~\ref{fig:contour_sample_frame}(a), the U-net yielded unrealistic trajectories that do not match with the left atrial dynamics, and the trajectories are corrected with the application of the UKF, as shown in Fig.~\ref{fig:contour_sample_frame}(b). The segmentation results by the proposed approach over different temporal frames for \emph{2ch}, \emph{3ch} and \emph{4ch} is given in Fig.~\ref{fig8}.

\subsubsection{Ablation study}
Table \ref{tab:without_classnetB} reports the results for U-net+UKF approach without \textit{Classnet-A} and \textit{Classnet-B} evaluated in terms of Dice score and Hausdorff distance. The reported values in the table demonstrate the impact of \textit{Classnet-B} in improving the segmentation accuracy by automatically selecting the images that require post-processing with UKF instead of applying it to the entire image set. The improvement was observed in all chamber views in terms of Dice coefficient as reported in Table \ref{tab:without_classnetB} as well as in column 3 of Tables \ref{tab8}, \ref{tab9}, and \ref{tab10}.

\begin{table*}[]
\caption{Evaluation of the automated segmentation results over 20 patients data sets using Dice coefficient and Hausdorff distance metrics. The results of the U-net and the proposed U-net+UKF methods were compared against expert manual ground truth produced by clinicians with and without the application of \textit{Classnet-A} neural net. The evaluations were performed over 1515 images obtained from the long-axis cine 2-chamber view. The chamber group consisted of 505 images.}
\setlength{\tabcolsep}{2.5pt}
\centering
\renewcommand{\arraystretch}{1.2}
\begin{tabular}{ccccc|cccc}
\toprule
\multirow{3}{*}{Method} & \multicolumn{4}{c|}{\textbf{2ch Dice Coefficient (\%)}}                                      & \multicolumn{4}{c}{\textbf{2ch Hausdorff Distance (mm)}}                                    \\ \cline{2-9} 
                        & \multicolumn{2}{c}{without \textit{Classnet-A}} & \multicolumn{2}{c|}{with \textit{Classnet-A}} & \multicolumn{2}{c}{without \textit{Classnet-A}} & \multicolumn{2}{c}{with \textit{Classnet-A}} \\ \cline{2-9} 
                        & U-net             & U-net+UKF         & U-net        & U-net+UKF           & U-net            & U-net+UKF          & U-net        & U-net+UKF           \\ \hline
\textbf{Train 20}                & $84.3\pm 6.7$         & $85.4\pm 6.0$          & $82.5\pm 13.5$    & $86.8\pm 9.7$            & $9.4\pm 5.4$          & $9.1\pm 5.0$            & $12.2\pm 8.5$     & $10.2\pm 7.7$            \\ \hline
\textbf{Train 40}                & $90.8\pm 4.0$          & $91.6\pm 3.2$          & $91.8\pm 7.6$     & $92.9\pm 4.2$            & $9.5\pm 6.9$          & $8.5\pm 5.0$            & $8.6\pm 6.8$      & $7.7\pm 6.2$             \\ \hline
\textbf{Train 60}                & $91.1\pm 3.8$          & $91.8\pm 3.0$          & $93.7\pm 2.6$     & $93.7\pm 2.7$            & $8.8\pm 6.9$          & $7.6\pm 4.6$            & $6.7\pm 3.5$      & $\mathbf{6.0\pm 2.9}$    \\ \hline
\textbf{Train 80}                & $93.4\pm 3.0$          & $93.5\pm 3.0$          & $93.5\pm 5.8$     & $\mathbf{94.1\pm 3.7}$   & $7.7\pm 6.4$          & $6.6\pm 4.3$            & $7.0\pm 4.3$      & $\mathbf{6.0\pm 3.3}$    \\ \bottomrule
\end{tabular}
\label{tab8}
\end{table*}

\begin{table*}[]
\caption{Evaluation of the automated segmentation results over 20 patients data sets using Dice coefficient and Hausdorff distance metrics from the long-axis cine 3-chamber view. The chamber group consisted of 505 images.}
\setlength{\tabcolsep}{2pt}
\centering
\renewcommand{\arraystretch}{1.2}
\begin{tabular}{ccccc|cccc}
\toprule
\multirow{3}{*}{Method} & \multicolumn{4}{c|}{\textbf{3ch Dice Coefficient (\%)}}                                      & \multicolumn{4}{c}{\textbf{3ch Hausdorff Distance (mm)}}                                    \\ \cline{2-9} 
                        & \multicolumn{2}{c}{without \textit{Classnet-A}} & \multicolumn{2}{c|}{with \textit{Classnet-A}} & \multicolumn{2}{c}{without \textit{Classnet-A}} & \multicolumn{2}{c}{with \textit{Classnet-A}} \\ \cline{2-9} 
                        & U-net             & U-net+UKF         & U-net        & U-net+UKF           & U-net            & U-net+UKF          & U-net        & U-net+UKF           \\ \hline
\textbf{Train 20}                & $83.0\pm 6.6$         & $84.8\pm 8.0$          & $77.0\pm 9.5$    & $80.4\pm 8.7$            & $19.7\pm 20.9$          & $17.1\pm 16.2$           & $11.2\pm 9.3$     & $10.5\pm 9.3$            \\ \hline
\textbf{Train 40}                & $89.3\pm 5.0$          & $90.1\pm 10.7$          & $79.1\pm 16.2$     & $83.4\pm 14.3$            & $17.0\pm 22.8$          & $13.9\pm 16.5$           & $16.6\pm 15.6$      & $9.4\pm 8.8$       \\ \hline
\textbf{Train 60}                & $90.9\pm 5.3$          & $91.3\pm 10.5$          & $92.5\pm 8.4$     & $92.6\pm 8.4$            & $13.9\pm 18.3$          & $12.4\pm 15.1$        & $10.4\pm 15.0$      & $7.4\pm 5.4$   \\ \hline
\textbf{Train 80}                & $92.4\pm 4.0$          & $92.5\pm 9.1$          & $90.3\pm 15.2$     & $\mathbf{93.7\pm 8.2}$   & $16.3\pm 22.9$          & $12.3\pm 14.8$        & $5.7\pm 3.2$      & $\mathbf{5.5\pm 3.0}$     \\ \bottomrule
\end{tabular}
\label{tab9}
\end{table*}

\begin{table*}[]
\caption{Evaluation of the automated segmentation results over 20 patients data sets using Dice coefficient and Hausdorff Distance metrics from the long-axis cine 4-chamber view. The chamber group consisted of 505 images.}
\setlength{\tabcolsep}{1.7pt}
\centering
\renewcommand{\arraystretch}{1.2}
\begin{tabular}{ccccc|cccc}
\toprule
\multirow{3}{*}{Method} & \multicolumn{4}{c|}{\textbf{4ch Dice Coefficient (\%)}}                                      & \multicolumn{4}{c}{\textbf{4ch Hausdorff Distance (mm)}}                                    \\ \cline{2-9} 
                        & \multicolumn{2}{c}{without \textit{Classnet-A}} & \multicolumn{2}{c|}{with \textit{Classnet-A}} & \multicolumn{2}{c}{without \textit{Classnet-A}} & \multicolumn{2}{c}{with \textit{Classnet-A}} \\ \cline{2-9} 
                        & U-net             & U-net+UKF         & U-net        & U-net+UKF           & U-net            & U-net+UKF          & U-net        & U-net+UKF           \\ \hline
\textbf{Train 20}                & $82.0\pm 10.1$         & $84.4\pm 5.1$          & $85.6\pm 5.6$    & $86.4\pm 5.6$            & $17.3\pm 12.8$          & $14.9\pm 10.4$            & $11.7\pm 5.7$     & $9.3\pm 5.1$            \\ \hline
\textbf{Train 40}                & $87.2\pm 12.6$          & $88.0\pm 8.1$          & $88.9\pm 5.9$     & $89.6\pm 5.1$            & $12.9\pm 11.8$          & $10.4\pm 7.8$            & $9.9\pm 11.7$      & $8.5\pm 8.1$             \\ \hline
\textbf{Train 60}                & $87.9\pm 11.2$          & $88.5\pm 10.1$          & $87.1\pm 16.5$     & $88.4\pm 13.4$            & $12.7\pm 11.8$          & $10.4\pm 7.9$            & $11.6\pm 10.8$      & $10.1\pm 10.1$    \\ \hline
\textbf{Train 80}                & $88.9\pm 11.3$          & $\mathbf{90.2\pm 4.2}$          & $88.9\pm 10.7$     & $90.1\pm 8.1$   & $11.8\pm 10.9$          & $10.0\pm 7.6$            & $12.2\pm 5.6$      & $\mathbf{8.4\pm 4.2}$    \\ \bottomrule
\end{tabular}
\label{tab10}
\end{table*}

\begin{table*}[htbp]
\caption{Evaluation of automated segmentation results over 20 patient data sets for the U-Net+UKF approach without \textit{Classnet-A} and \textit{Classnet-B} in terms of Dice coefficient and Hausdorff distance metrics for the long-axis cine 2, 3, 4-chamber views. The test set for each chamber group consisted of 505 images.}
\setlength{\tabcolsep}{4.5pt}
\centering
\renewcommand{\arraystretch}{1.2}
\begin{tabular}{cccc|ccc}
\toprule
\multirow{2}{*}{Method} & \multicolumn{3}{c|}{\textbf{Dice Coefficient (\%)}}                                      & \multicolumn{3}{c}{\textbf{Hausdorff Distance (mm)}}                                    \\ \cline{2-7} 
                        & 2ch            & 3ch        & 4ch        & 2ch          & 3ch           & 4ch          \\ \hline
\textbf{Train 20}                & $84.6\pm 6.5$          & $83.3\pm 6.8$          & $83.5\pm 8.4$    & $9.5\pm 3.3$            & $9.6\pm 2.8$          & $9.7\pm 3.8$           \\ \hline
\textbf{Train 40}                & $90.9\pm 4.0$          & $89.4\pm 4.9$          & $87.7\pm 10.9$     & $7.5\pm 2.4$            & $10.0\pm 7.3$          & $9.7\pm 7.5$      \\ \hline
\textbf{Train 60}                & $91.2\pm 3.8$          & $91.0\pm 5.2$          & $88.2\pm 10.6$     & $7.0\pm 2.5$            & $8.6\pm 3.5$          & $10.0\pm 7.8$        \\ \hline
\textbf{Train 80}                & $93.5\pm 2.9$          & $92.5\pm 4.0$          & $89.9\pm 9.9$     & $7.0\pm 2.6$   & $9.1\pm 3.2$          & $8.5\pm 6.1$        \\ \bottomrule
\end{tabular}
\label{tab:without_classnetB}
\end{table*}

\begin{table*}[htbp]
\caption{The number of images utilized in training the U-Net with and without \textit{Classnet-A} for 2, 3, 4-chamber sequences. The image sequences have either 25 or 30 images leading to slight variations in the number of images  for 2, 3, and 4 chamber training sets.}
\centering
\begin{tabular}{cccc|c}
\toprule
\multirow{2}{*}{Method} & \multicolumn{3}{c|}{\textbf{With \textit{Classnet-A}}}   & \multicolumn{1}{c}{\textbf{Without \textit{Classnet-A}}}   \\ \cline{2-5} 
                        & 2ch            & 3ch        & 4ch       & Combined 2,3,4ch          \\ \hline
\textbf{Train 20}                & $500$          & $505$         & $505$        & $1510$          \\ \hline
\textbf{Train 40}                & $1020$         & $1025$        & $1025$       & $3070$        \\ \hline
\textbf{Train 60}                & $1535$         & $1540$        & $1540$       & $4615$        \\ \hline
\textbf{Train 80}                & $2115$         & $2015$        & $2090$       & $6220$         \\ \bottomrule
\end{tabular}
\label{tab:num_samples}
\end{table*}
\section{Discussion}
In this study, we proposed and validated a fully automated segmentation approach to delineate left atrium from MRI sequences using deep convolutional neural networks and Bayesian filtering. Previous automated left atrial segmentation method attempted to delineate the chamber from a single time point in the cardiac phase using static images \citep{hunold2003radiation,ordas2007statistical,ecabert2008automatic, ecabert2011segmentation,zhu2012automatic, mortazi2017cardiacnet, pop2019statistical}. The proposed study relied on 2-chamber, 3-chamber and 4-chamber 2D MRI sequences, which may or may not be orthogonal to each other. These long-axis sequences do not form a 3D MRI dataset, as in the case of STACOM MICCAI 2018 challenge \citet{pop2019statistical}, however, offer a series of images to assess the temporal characteristics of the left atrium over the cardiac cycle. To the best of authors' knowledge, this is the first study to propose a fully automated approach to delineate left atrium over the entire cardiac cycle that could serve as the basis for the functional assessment of the chamber.

The intent of \emph{Classnet-A} is to automatically identify the 2-chamber, 3-chamber and 4-chamber views using image pixel information only. The experimental analysis showed that the \emph{Classnet-A} achieved perfect classification performance without a single misclassification when trained with images from 80 patients. The results presented in Tables \ref{tab8}, \ref{tab9}, and \ref{tab10} show that the application of \emph{Classnet-A} led to an improvement on segmentation accuracy measured in terms of Dice score in seven cases, slight reduction ($<1\%$) or similar performance in four cases, and a more considerable reduction in one case for Train 60 and Train 80 scenarios where the algorithm yielded a perfect classification as indicated in Table \ref{tab7}. The impact of \textit{Classnet-A} on the overall performance becomes unpredictable with smaller training sets such as \textit{Train 20} scenario where \textit{Classnet-A} led to inaccurate classifications in addition to the varying performances of the corresponding U-Net algorithms. Although the U-Net with \textit{Classnet-A} led to improved performance over the U-Net without \textit{Classnet-A} for \textit{Train 20} scenario for 4-chamber sequences as reported in Table \ref{tab10}, no similar improvements were observed for 2 and 3 chamber sequences as reported in tables \ref{tab8} and \ref{tab9}. Future studies will explore the performance of neural networks with smaller training sets.

The intent of \emph{Classnet-B} is to detect images that will lead to less accurate left atrial segmentation by U-net, and therefore, the training phase of \emph{Classnet-B} utilized the ground truth data from the training set. Upon trained, the \emph{Classnet-B} neural net parameters were no longer updated and none of the ground truth delineations from the test set was utilized in the process. 

Among the three long-axis sequences, the 3-chamber view requires more experience to segment as the anterior margin of the mitral valve, being adjacent to the aortic valve, is harder to delineate than in other views. The results of this study show that the performance is lower for the 3-chamber view in comparison to other views when the proposed approach was trained with images from 20 patients. However, as indicated in Table \ref{tab9}, a significant performance improvement was achieved in 3-chamber view results when the training set was increased to 80 patients.

In this study, the test set MRI sequences were selected from 20 patients, and none of the scans were excluded in the analysis regardless of the image quality, which led to high inhomogeneity among the test images and a high standard deviation was observed for the evaluation metrics reported in Tables \ref{tab8}, \ref{tab9}, and \ref{tab10}. Table \ref{tab:num_samples} reports the numbers of samples used for training the U-Net approach with and without \textit{Classnet-A}. Due to the different number of samples used for training the U-Net approach, the performance of the algorithm varies even when the \textit{Classnet-A} offers perfect classifications. For instance, the results for the \textit{Train 80} scenario reported in columns 3 and 5 for the U-Net approach with and without \textit{Classnet-A} in Tables \ref{tab8}, \ref{tab9}, \ref{tab10} are different where Classnet-A yielded perfect classification.

Clinical importance of the left atrial functional analysis has been emphasized in many echocardiography studies particularly the prognostic value of left atrial measurements for patients with heart failure with preserved ejection fraction \citep{Morris2011,Santos2014,Santos2016,Cameli2017,Liu2018}. Clinical studies that used MRI sequences for cardiac functional measurements also led to similar findings and demonstrated the prognostic values of the left atrial functional parameters in heart failure patients \citep{VonRoeder2017,Chirinos2018}. One of the significant advantages of the proposed method is that it is fully automated and, therefore, will allow for analyzing a large number of MRI images which otherwise tedious and time-consuming. Further clinical studies are warranted to objectively measure the left atrial function and assess its predictive ability for heart failures with preserved and reduced ejection fraction. 

The proposed approach performs semantic segmentation to delineate left atrium where pixels are classified either foreground or background. Although such delineations could be used for computing functional parameters such as volume or volume rate  over time, some additional parameters such as strain or strain-rate require open contours representing the boundary of the chamber walls that excludes anatomical structures corresponds to the mitral valve. Future studies will attempt to address this limitation.

\section{Conclusion}
This paper proposes a fully automated approach to segment the left atrium from routine long-axis cardiac MR image sequences acquired over the entire cardiac cycle. The proposed methods introduces a neural net approach to classify the input images and loads the corresponding U-net based convolutional neural network architecture for semantic segmentation. A second classifier network is developed to select sequences for further processing with the Unscented Kalman filter to impose temporal periodicity over the cardiac cycle. The proposed algorithm was trained separately with different scenarios with varying amount of training data with images acquired from 20, 40, 60, 80 patients. The prediction results were compared against expert manual delineations over an additional 20 patients in terms of the Dice coefficient and Hausdorff distance. The quantitative assessment showed that the proposed method performed better than recent U-net algorithm. The proposed approach yielded $100\%$ accuracy for chamber classification and a mean Dice coefficient of $94.1\%$, $93.7\%$ and $90.1\%$ for 2, 3 and 4-chamber sequences after training using 80 patients data.

\section*{Acknowledgments}
X. Zhang was supported by the Mitacs Globalink Research Internship program. D. Glynn Martin was supported by the University of Alberta Radiology Endowed Fund. The work in-part was supported by a grant from the Servier Canada Inc.

\balance
\bibliographystyle{model2-names}\biboptions{authoryear}
\bibliography{media-revision3_preprint}

\end{document}